# A Literature Review on Length of Stay Prediction for Stroke Patients using Machine Learning and Statistical Approaches

Ola Alkhatib[1] and Ayman Alahmar[2]

[1]*Department of Computer Science, Lakehead University, Thunder Bay, Ontario, Canada*
[2]*Department of Software Engineering, Lakehead University, Thunder Bay, Ontario, Canada*

**Abstract** Hospital length of stay (LOS) is one of the most essential healthcare metrics that reflects the hospital quality of service and helps improve hospital scheduling and management. LOS prediction helps in cost management because patients who remain in hospitals usually do so in hospital units where resources are severely limited. In this study, we reviewed papers on LOS prediction using machine learning and statistical approaches. Our literature review considers research studies that focus on LOS prediction for stroke patients. Some of the surveyed studies revealed that authors reached contradicting conclusions. For example, the age of the patient was considered an important predictor of LOS for stroke patients in some studies, while other studies concluded that age was not a significant factor. Therefore, additional research is required in this domain to further understand the predictors of LOS for stroke patients.

*Keywords: Length of Stay, Stroke, Machine Learning, Data Mining, Statistical Analysis.*

## 1. INTRODUCTION

HEALTHCARE sectors show increasing costs in most regions around the world. Healthcare expenditure constitutes a significant share of the gross domestic product for many countries. There are many challenges associated with growth in the healthcare sector, including increased pressure on the limited resources of hospitals. This issue has motivated researchers to conduct further research related to hospital resource optimization. Since hospitalization constitutes a significant cost of patient care, many researchers have been investigating the problem of patient Length of Stay (LOS) prediction. LOS is defined as the duration of a patient hospitalization, and it is determined as the difference between the timestamp of a patient hospital discharge and the timestamp of their hospital admission [1], [2]. LOS prediction is an important topic for many reasons, such as:

- Knowledge of LOS allows hospitals to manage their bed and room capacities so that they can know how long a patient is expected to occupy hospital space.
- Information about LOS allows hospitals to determine the number of staff that must be scheduled over the day/night shifts to properly accommodate the patients.
- Patients and their families can estimate the cost of a stay in paid hospitalizations.

By investigating the existing literature, we found that LOS prediction papers used machine learning/statistical approaches and can be divided into the type of disease under consideration. Some authors studied LOS prediction in general (i.e. without specifying a specific disease), while other researchers focused on LOS prediction pertinent to a specific disease (e.g., stroke, diabetes). Fig. 1 depicts this categorization and includes example references to research articles.

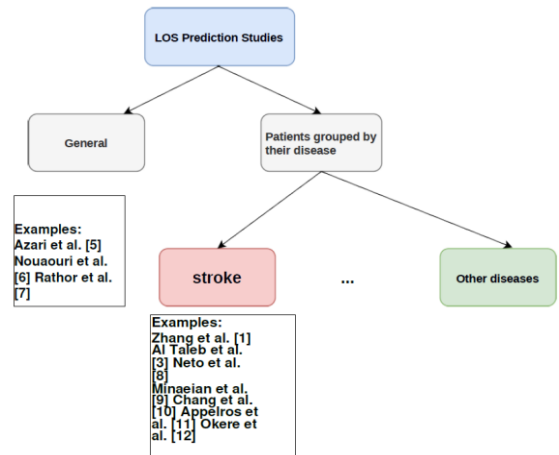

**Fig. 1.** Classification of LOS prediction research articles.

In this survey, we review papers that predict the patients' LOS in general and then we focus on LOS prediction for stroke patients. We focused on stroke patients because they face many challenges with definite need for hospitalization, and also because strokes have an enormous cost on healthcare systems around the world. Using popular research search engines (e.g., IEEE Xplore, Springer, Science Direct, etc.), we searched phrases such as "length of stay", "hospital length of stay", "prediction", "machine learning", "stroke", "ischemic stroke", "data mining", and "statistical analysis" in order to find existing work on this topic (up to mid 2020).

Stroke is a disease that affects the arteries leading to and within the brain. It can be a significant financial and health burden for patients, medical staff, and healthcare systems. Stroke is associated with prolonged LOS in hospitals and rehabilitation facilities [1] and is a leading cause of death and disability worldwide. According to Statistics Canada, in 2018, stroke was the third largest cause of death in Canada after cancer and heart disease (https://www.statcan.gc.ca). Stroke, also known as cerebrovascular accident (or CVA), is a sudden and devastating illness that is characterized by the rapid loss of the functions of the brain due to a disruption of blood flow to the brain (see Fig. 2). This disruption is caused by: lack of blood flow (ischemic strokes), which account for more than 80% of all strokes; blockage of blood flow; or hemorrhage [3].

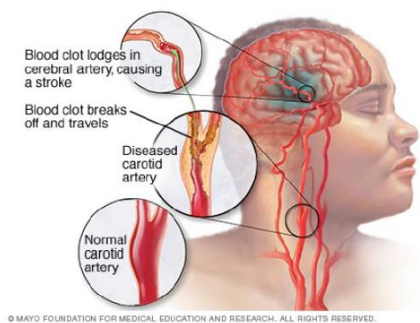

**Fig. 2.** Illustration of stroke (Source: Mayo Clinic (mayoclinic.org)).

The remainder of the paper is organized as follows. Below we present our literature review on LOS prediction for general patients. Next we consider stroke patients. Finally, the paper ends with a discussion and conclusions section.

## 2. LOS PREDICTION FOR GENERAL PATIENTS

Kabir et al. [4] proposed a non-linear feature selection method using artificial neural networks (ANNs) to determine the most essential features for LOS prediction. The study evaluates the performance of (ANNs), support vector machines (SVM), and logistic regression (LR) on selected subsets of features to predict the LOS class and identify the best subset of features. The study used a dataset from the National Surgical Quality Improvement Program database. The dataset, based on data from 2015, included 273 features of more than 880,000 surgical patients admitted to hospital for various surgical procedures to address different diseases and medical conditions. The authors reduced the features from 273 to 40 based on consultation with domain experts (anesthesiologists). After preprocessing the dataset, 715,143 patient records were selected for further analysis. The patients were categorized using their medical group into the following nine surgical categories: (1) general surgery, (2) vascular, (3) urology, (4) plastics, (5) otolaryngology, (6) orthopedics, (7) gynecology, (8) neurosurgery, and (9) other surgical conditions (including thoracic, cardiac surgery, and interventional radiology patients). An additional category that includes all patients was added as group (10) for comparison purposes. The authors presented the normalized importance score of features for each patient category. In general, features 16, 20, and 21 showed the most significant contribution to the prediction of LOS. Feature 16 indicates the period from admission to surgery, feature 20 denotes whether the patient is an outpatient or inpatient, and feature 21 represents the estimated probability of morbidity computed by the hospital using LR. These features had a strong correlation with LOS which validates the performance of the non-linear approach of this study in obtaining considerable features. Another important factor presented in this research is the specific correlation of features with their categories. For example, feature 38, which represents the situation of a patient's wound, was important in predicting the LOS for otolaryngology patients, while it had less importance in LOS predicting for other patient categories. By comparing the importance of features computed for all patients (group 10) with other categories, the authors showed that grouping patients based on their disease can improve the accuracy of LOS predictive models. The final results revealed that ANNs, as non-linear classifiers, beat SVMs and LR in the achieved accuracy for LOS prediction. This proves that the relationship between LOS and its predictors is highly non-linear. The ANNs model improves accuracy and eliminates the number of required features.

Azari et al. [5] used a multi-tiered data mining approach for predicting hospital LOS to decrease the uncertainty related with the LOS for inpatients. Their prediction approach was based on clustering (i.e., k-means clustering) to create the training sets that train various classification algorithms. The number of clusters was determined based on the disease conditions or by using the Charlson index which provides the general categories of the diseases. Fig. 3 describes the approach used in this study. Several classifiers were used to predict the LOS such as: K-nearest neighbors, LR, naive Bayes, SVMs, Bayesian networks (Bnets), J48 decision tree, classification rules (JRip), bagging, random forest, and boosting. The paper considered various performance metrics such as accuracy, Kappa statistic, precision, recall, and area under the curve (AUC). In order to rank the classifiers, researchers used the Friedman test to determine the classifier with the best outcome for a certain level of clustering. The dataset used the Heritage Health prize data. The dataset contains 1,048,576 records of hospital claims within a 3-year period. The results showed that using clustering as a precursor to form the training set provides better results compared to non-clustering based training sets. The results also showed that Bnets, SVMs, JRip, Bagging, and J48 had better overall performance than the other classifiers. The outcomes of the paper were validated by a domain expert from Emergency Medicine.

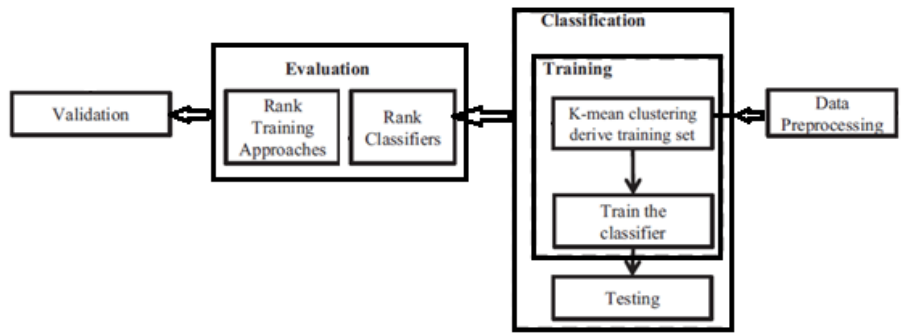

**Fig. 3.** LOS prediction approach of Azari et al. [5].

Nouaouri et al. [6] proposed the application of data mining techniques to predict the LOS for patients without considering a specific disease. They introduced the Evidential LOS prediction Algorithm (ELOSA) that allows the prediction of the LOS of a new patient. Their approach handles the imprecision, uncertainty, and missing data within the dataset of patients. The ELOSA algorithm is based on the precise support and association rule confidence measures [6]. The LOS experiments were conducted on a real hospital dataset that contains the data for 270 patients. To predict the inpatient LOS, the authors considered age, sex, physiological conditions (emergency degree), and operation length. The emergency degree column in the dataset was a categorical attribute that contained values from the following set {A, R, D}. A indicates an absolute emergency, R is for relative emergency, and D represents delayed emergency. The length of stay was classified as Short (S), approximately 3 days, Medium (M), approximately 10 days, and Long (L), approximately 20 days, with a small overlap between the ranges as shown in Fig. 4. They compared their results with other algorithms in similar studies and concluded that their approach showed better results.

Rathor et al. [7] used a clustering algorithm (i.e., Density Based Spatial Clustering of Applications with Noise (DB- SCAN)) and K-Apriori, which is a combination of Apriori and K-means algorithms. The algorithms were applied to a dataset of 9,052 patients (the source of the dataset was not disclosed). The execution time of the algorithms were compared, showing DBSCAN to be faster than the K-Apriori; although, DBSCAN took exponentially longer as the number of inputs increased. The prediction was based on the current symptoms and medical history of the patients, which was provided by the patient at the time of admission. For prediction of LOS, the medical data underwent a pre-processing phase, which had three steps:

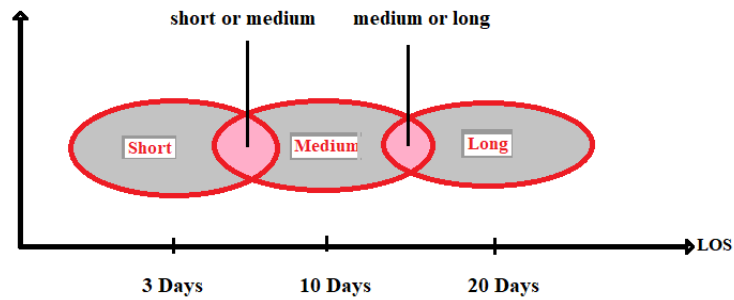

**Fig. 4.** LOS classes in terms of days [6].

data cleaning, data integration and transformation, and data reduction. Then, using the processed data, symptoms for a particular disease were grouped together and used for LOS prediction. The study determined the times of execution of K-apriori and DBSCAN independently and subsequently compared them. Both algorithms were treated with the same number of inputs and with the same values. The authors concluded that the execution time of DBSCAN was comparatively much shorter than K-Apriori, but as the number of inputs increase to high values, the execution time of DBSCAN increased exponentially whereas there was no change in K-Apriori.

### 3. LOS PREDICTION FOR STROKE PATIENTS

In stroke, the brain is prevented from getting oxygen and nutrients from the blood. Without oxygen and nutrients, brain cells begin to die within minutes. Sudden bleeding in the brain can also cause a stroke if it damages brain cells. A stroke is a medical emergency that can cause lasting brain damage, long-term disability, or even death. Signs of a stroke can range from mild weakness to paralysis or numbness on one side of the face or body. Other signs include a sudden and severe headache, sudden weakness, trouble seeing, and trouble speaking or understanding speech (https://www.nhlbi.nih.gov/health-topics/stroke).

Al Taleb et al. [3] introduced a machine learning method for early prediction of LOS of stroke patients. They tested their approach at the Stroke Unit of King Fahad Bin Abdul- Aziz Hospital in Saudi Arabia. The study was based on 866 stroke patients, whose data was retrieved from the Neurology Department database. For data cleaning, each set of patient data was manually examined for invalid or erroneous inputs. Records with missing values in more than 50% of the attributes were deleted. For the records with missing values in less than 50% of attributes, missing values were replaced with the average value of the respective attributes for numeric attributes, and with the mode value for the categorical attributes. The approach involved a feature selection step based on Information Gain (IG) followed by a prediction model development step using different machine learning algorithms as explained below.

The original dataset contained 105 attributes, out of which 54 attributes were manually eliminated due to being irrelevant or redundant, such as time of arrival, date of MRI, and cause of death. The remaining 51 attributes were ranked based on their IG with respect to LOS, and then an iterative process of elimination was applied where the researchers began processing all of the features. Then features were eliminated one at a time, starting with the least ranked one, and the IG was recalculated. The repetitive process stopped when there was no further improvement in IG. Finally, 16 remaining attributes (including LOS) were selected for the prediction steps. The selected attributes and their IG values are listed in Fig. 5.

| Attributes (abbreviated names) | Information Gain |
|---|---|
| NIHSS Level (NHS) | 0.12628 |
| TOAST classification (TST) | 0.11896 |
| NGT (NGT) | 0.09453 |
| Dysphagia Screen Result (DSR) | 0.07997 |
| Mode of Arrival (MA) | 0.05825 |
| Pneumonia (PN) | 0.05772 |
| Sensory changes (SC) | 0.05108 |
| UTI (UTI) | 0.05099 |
| Age group (AGE) | 0.04676 |
| MRS level (MRS) | 0.04669 |
| Brathel Index (BI) | 0.04429 |
| Motor (weakness, MOT) | 0.02948 |
| LDL Level (LDL) | 0.02811 |
| Plavix (PLV) | 0.02674 |
| Hypertension (HYP) | 0.02409 |
| Length of Stay (class attribute, LOS) | |

**Fig. 5.** Selected attributes and their IG values with respect to the class attribute (LOS) [3].

Prediction results were compared to identify the algorithm with the best performance. Several experiments were performed in various settings. The authors found that the most accurate model in their study was the Bnet model with accuracy of 81.28%.

In another study, Neto et al. [8] proposed a neural network LOS prediction method based on the information available on the stroke neurological events, the patient's health status, and surgery details. The neural network was trained to test with three attribute subsets of different sizes. The first subset contained 33 attributes, the second 14, and the third subset consisted of only 7 attributes. By testing the three subsets, it was possible to define an optimal neural network configuration where the lowest error values were registered as Root Mean Squared Error, 5.9451, and Mean Absolute Error, 4.6354. They concluded that the third use case (the one with fewer variables) obtained better results than the other attribute sets.

Zhang et al. [1] aimed to develop a risk prediction model of prolonged LOS in stroke patients for 50 inpatient rehabilitation centers in 20 provinces across mainland China, based on the International Classification of Functioning, Disability, and Health Generic Set case mix on admission. The study was conducted on 383 stroke patients. The independent predictors of prolonged LOS were identified using Multivariate Logistic Regression (MLR) analysis. A prediction model was established and then evaluated by receiver operating characteristic curve analysis, and the Hosmer-Lemeshow test. The results showed that the type of medical insurance and the performance of daily activities were associated with prolonged LOS. Age and mobility level demonstrated no significant predictive value. The prediction model revealed acceptable discrimination shown by an AUC of 0.699. The researchers concluded that the scores for the type of medical insurance and the performance of daily activities on admission were independent predictors of prolonged LOS for stroke patients. Their study proved that prediction models allow stakeholders to quantitatively estimate the risk of prolonged LOS upon admission, and to facilitate financial planning. They can also determine any required treatment regimens during hospitalization, the need for referral after discharge, and reimbursement of costs.

Minaeian et al. [9] sought to determine whether a longer emergency LOS was associated with a poor 90-day outcome following an ischemic stroke. Their method was based on a retrospective analysis of a single-center cohort of consecutive ischemic stroke patients. There were 325 patients in the study. They constructed multivariable linear and LR models to determine factors independently associated with emergency LOS as well as a poor 90-day outcome. The results revealed that the median LOS in the cohort was 5.8 hours of time spent in emergency. For patients admitted to the inpatient stroke ward (160 patients) versus neurointensive care unit (NICU) (165 patients), the median LOS was 8.2 hours versus 3.7 hours, respectively. On multivariable linear regression, NICU admission, endovascular stroke therapy, and thrombolysis were inversely associated with the LOS. Evening shift presentation was associated with a longer LOS. On MLR, a greater admission stroke severity, worse pre-admission modified Rankin scale, hemorrhagic conversion, and a shorter LOS were associated with a poor 90-day outcome. Early initiation of statin therapy, endovascular stroke therapy, NICU admission, and evening shift presentation were associated with a good 90-day outcome. The authors stressed in their conclusion that in contrast to prior studies, a shorter emergency LOS was associated with a worse 90-day functional outcome, possibly reflecting prioritized admission of more severely affected stroke patients who were at high risk for a poor functional outcome.

Chang et al. [10] aimed to determine the clinical and demographic predictors of LOS of acute care hospital stay for patients with first-ever ischemic stroke. In the study, a group of 330 patients who had their first-ever ischemic stroke and were admitted to a medical center in southern Taiwan were followed prospectively. The researchers evaluated only the factors that could be known at the time of admission. Univariate analysis and multiple regression analysis were used to identify the LOS main predictors.

In the reported results, the median LOS was 7 days, average LOS was 11 days, and the LOS range was 1 to 122 days. Among the prespecified demographic and clinical characteristics, the National Institutes of Health Stroke Scale (NIHSS) score at admission, the quadratic term of the initial NIHSS score, the modified Barthel Index score at admission, small-vessel occlusion stroke, smoking, and sex were the main predictors for LOS. In particular, for each 1-point increase in the score of NIHSS, LOS increased by approximately 1 day for patients with mild or moderate neurological impairments (score 0 to 15 points), while LOS decreased approximately 1 day for patients with severe neurological impairments (score 15 points). The authors concluded that the severity of acute stroke, as scored by the total score on NIHSS, was an important factor influencing LOS after acute stroke hospitalization.

Appelros et al. [11] examined the factors that influence acute and total LOS for stroke patients. The basis of their investigation was a population-based cohort of first-ever stroke patients (388 patients). Patient data included age, sex, risk factors, social factors, dementia, stroke type, and stroke severity, measured using the NIHSS. The results showed a mean acute LOS of 12 days and mean total LOS of 29 days. Independent predictors of acute LOS were stroke severity, lacunar stroke, pre-stroke dementia, and smoking. Independent predictors of total LOS were stroke severity and pre-stroke activities of daily living dependency. The NIHSS elements that best correlated with LOS included paresis, unilateral neglect, and level of consciousness. The conclusion was that stroke severity is a strong and reliable predictor of LOS. The results can be used as a baseline for evaluating cost-effectiveness of stroke care changes, such as assessment of new drugs and organizational modifications.

The study conducted by Okere et al. [12] was designed to evaluate predictors of hospital LOS and re-admissions among non-surgical ischemic stroke patients. The patients in this study were adult patients ($\geq$ 18 years) with a diagnosis of non-surgical ischemic stroke, who were hospitalized between November 2007 and March 2013. The results of the statistical analyses (multivariate and bivariate analyses), revealed that insurance type was a significant predictor of LOS, with Medicare patients having a longer LOS compared to patients with private insurance. Severity of illness was also a predictor of LOS, whereby patients prescribed statins and patients aged less than 80 years old had a lower 30-day hospital re-admission rate compared to patients who were not prescribed statins and who were older than 80 years of age, respectively.

Choi et al. [13] considered LOS prediction for acute stroke patients and extracted their dataset from 2013 and 2014 discharge injured patient data. The data was classified as 60% for training and 40% for evaluation. In their model, they used the multiple regression analysis method combined with machine learning techniques (such as decision tree and neural network) to create an ensemble technique that integrates all methods. They evaluated their model using root absolute error index. Considering the used methods, the error index was 23.7 for multiple regression, 23.7 for decision tree, 22.7 for neural network, and 22.7 for the ensemble technique. They concluded that the neural network technique was found to be superior (even reaching the level of ensemble methods).

In the study carried out by Svendsen et al. [14], the author's objective was to determine whether healthcare quality was associated with LOS among stroke patients. They performed a population-based study that included 2,636 stroke patients between 2003 and 2005 from a stroke unit in Denmark. In this study, quality of care was measured as fulfillment of twelve (12) criteria: "early admission to a stroke unit, early antiplatelet therapy, early anticoagulant therapy, early computed tomography/magnetic resonance imaging scan, early water swallowing test, early mobilization, early intermittent catheterization, early deep venous thromboembolism prophylaxis, early assessment by a Physiotherapist and an Occupational Therapist, and early assessment of nutritional and constipation risk" [14]. The authors' analyzed the patients' data using linear regression clustered at the stroke units by multilevel modeling. The results showed that the median length of stay was 13 days. Fulfilling each quality of care criterion was associated with shorter LOS. The authors found that "the association between meeting more quality of care criteria and LOS followed a dose-response effect, that is, patients who fulfilled between 75% and 100% of the quality of care criteria were hospitalized only one-half as long as patients who fulfilled between 0% and 24% of the criteria". The study concluded that the care in the early phase of stroke is very important as a high initial quality of care was associated with shorter length of stay among stroke patients.

Garza-Ulloa [15] used neural network algorithms to predict rehabilitation LOS for stroke patients along with other stroke metrics (i.e., the need for surgery and rehabilitation need). The study objective was to find an optimal neural network configuration using three different available software: one manual (with no automatic stepwise functions and limited diagnostic capability), another semi-automatic (allows step- wise function with good diagnostics), and neuro-intelligence (uses genetic algorithm to find the best neural network (NN) configuration).

Based on the 14 stroke input variables and the 3 output target stroke values, the paper suggested that the forecasting of: surgery, rehab and days of rehabilitation were possible using neural network tools. Fig. 6 (from [15]) outlines the 14-group variables for the proposed neural network .

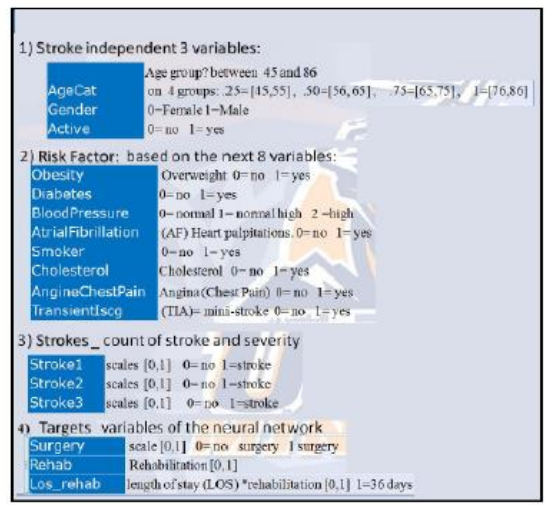

**Fig. 6.** 14-Group variables for the proposed NN [15].

The study of Ng et al. [16] aimed to investigate LOS characteristics and identify the predictors of post-stroke acute, rehabilitation and total LOS. The study divided the stroke patients (1,277 patients) into two subgroups of short LOS and long LOS, and compared the two subgroups regarding complication rates and functional outcomes. The authors considered stroke patients within a 5-year period from 2004 to 2009 in a dedicated rehabilitation unit within a tertiary academic acute hospital in Singapore. The primary outcome measure considered in the rehabilitation phase was the functional independence measure (FIM). Short acute LOS patients were defined as patients who stayed less than 7 days. Most patients in the study were ischemic stroke patients (1,019 patients (80%)), while the remaining patients were haemorrhagic stroke patients (20%). The results of the study showed that the average acute and rehabilitation LOS were 9-7 days and 18-10 days, respectively. "Haemorrhagic strokes and anterior circulation infarcts had significantly longer acute, rehabilitation and total LOS compared to posterior circulation and lacunar infarcts" [16]. Patients that were admitted after 2007 had significantly shorter acute, rehabilitation and total LOS. The authors found poor correlation between the acute and rehabilitation LOS (r = 0.12). In multivariate analysis, considering rehabilitation LOS, admission FIM scores were significantly associated with LOS, while, in acute LOS, stroke type was strongly associated with LOS. "Patients in the short acute LOS group had fewer medical complications and similar FIM efficacies compared to the longer acute LOS group." [16]. The authors concluded that it is very important to transfer appropriate patients as early as possible to rehabilitation units as this ensures that the development of clinical complications is minimized, while rehabilitation efficacy is maintained.

Bindawas et al. [17] aimed to investigate the association between LOS and functional outcomes among patients with stroke discharged from a rehabilitation facility in Saudi Arabia. There were 409 adult patients in the study (age 18) admitted between 2008 and 2014, with no deaths during the study period. Patients were divided into 4 different groups based on the days of rehabilitation: ≤ 30 days (n=114), 31–60 days (n=199), 61–90 days (n=72), and > 90 days (n=24). Multivariate regression analyses were used to evaluate functional outcomes using the FIM. The results of the study showed that higher FIM scores were significantly associated with a LOS ≤ 30 days and 31–60 days, compared to > 90 days. The authors concluded that "a short or intermediate LOS is not necessarily associated with worse outcomes, assuming adequate care is provided." [17].

Arboix et al. [18] considered the identification of clinical predictors of prolonged hospital stay after acute stroke. They considered a long study period of 17 years for patients in Spain who have had their first-ever ischemic stroke and primary intracerebral hemorrhage. Prolonged LOS stay was defined as LOS longer than 12 days after admission. The attributes considered included demographic data, cardiovascular risk factors, neuroimaging findings, clinical factors, and outcome. LR analysis was used to evaluate the independent influence of statistically significant variables in the duration of hospitalization. The results of 3,112 acute stroke patients showed that prolonged hospital stay was recorded in 1,536 (49.4%) cases. Furthermore, males, limb weakness, vascular complications, urinary complications, and infectious complications were independently associated with longer LOS, whereas being symptom free at hospital discharge and lacunar infarction were inversely associated with prolonged LOS. The authors concluded that "in-hospital medical complications (vascular, urinary, and infectious) are relevant factors influencing duration of hospitalization after acute stroke. Therefore, prevention of potentially modifiable risk factors for medical complications is an important aspect of the early management of patients who experienced stroke" [18].

The objective of the study conducted by Koton et al. [19] was to derive a simple score for the assessment of the risk of prolonged length of stay for acute stroke patients. Prolonged LOS was defined as LOS ≥ 7 days. The results showed that the severity of stroke was the strongest multivariable predictor of prolonged LOS. The study concluded that a simple prolonged LOS score, based on available baseline information (stroke severity), may be useful for developing policies aimed at better use of resources and optimal discharge planning of acute stroke patients [19].

Hung et al. [20] aimed to consider the factors that influence LOS for stroke patients in Taiwan. The researchers explored how intravenous thrombolysis (IVT) affects LOS in an acute care hospital setting. The study considered adult patients with ischemic stroke who presented within 48 hours of stroke onset. The relationship between IVT and prolonged length of stay (LOS ≥ 7 days) was studied by both classification and regression tree, as well as MLR analyses. Fig. 7 illustrates the risk stratification for prolonged LOS by means of the classification and regression tree analysis. Among the study population of 3,054 patients, 1,110 presented within 4.5 hours. The median LOS was 7 days (ranging from 4 to 11 days), and 1,619 patients had prolonged LOS. MLR revealed that IVT was an independent factor that reduced the risk of prolonged LOS, whereas age, NIHSS score, diabetes mellitus, and leukocytosis at admission predicted prolonged LOS. Decision tree analysis identified four variables (NIHSS score, IVT, leukocytosis at admission, and age) as important factors and they were used to partition the patients into six subgroups (see Fig. 7). The patient subgroup that had an NIHSS score of 5 to 7 and received IVT had the lowest probability (19%) of prolonged LOS [20]. The authors concluded that IVT minimized the risk of prolonged length of stay in patients with acute ischemic stroke. They recommended that measures to increase the rate of IVT be encouraged.

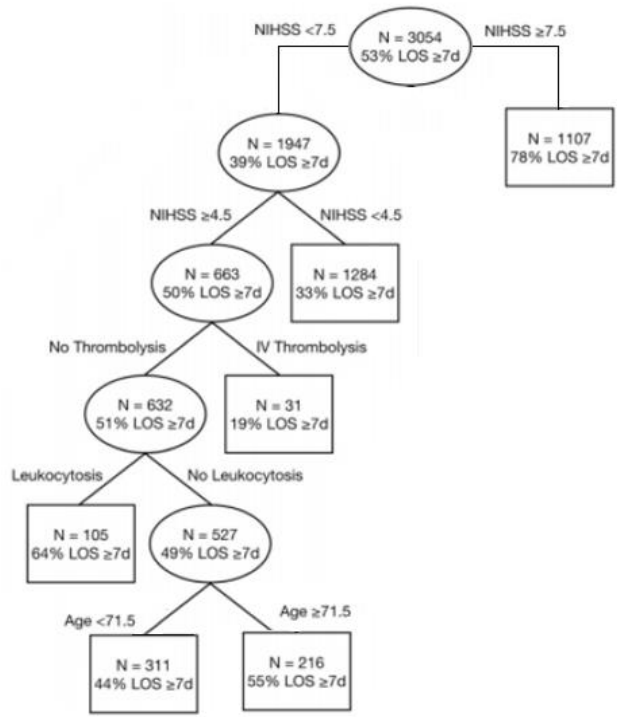

**Fig. 7.** Risk stratification for prolonged LOS by means of the classification and regression tree analysis [20].

## 4. DISCUSSION AND CONCLUSIONS

The topic of hospital LOS is an important topic for hospital resource utilization and optimization [21][22]. Although many researchers have conducted research on LOS prediction, more research is needed to further enrich this domain of study. We noticed from our literature review that researchers occasionally reached contradicting conclusions about LOS for stroke patients. For example, some researchers (e.g., [1]) found that the age of the patient was not a significant predictor of prolonged LOS, whereas, others (e.g., [20]) concluded that a patient's age was an important predictor for LOS for stroke patients. This shows how LOS prediction is a complex phenomenon that requires more studies and careful investigation. On the other hand, most researchers (implicitly or explicitly) agreed that stroke severity (e.g., NIHSS) is a major predictor of LOS in stroke patients.

Another observation from our literature review is that not all researchers teamed up with domain experts while conducting machine learning studies on LOS prediction. We view this as a limitation in such studies because LOS prediction is a topic that is highly related to the medical field as well as to patients' medical data and characteristics. This implies that in future research, domain experts should be consulted before finalizing and publishing such LOS prediction studies. Domain experts enrich the studies and make them more realistic.

Another point is that some researchers did not specifically mention the attributes that were effective in LOS prediction at the end of their studies. However, this particular information is highly important for the readers of such research articles so that new research can build on previous studies. We also recommend more future cooperation between machine learning researchers in this field because of its importance in hospital resource utilization, decreasing healthcare costs, and achieving healthier people and an overall healthier society.